\documentclass[sigconf,natbib=false]{acmart}
\DeclareUnicodeCharacter{0301}{}
\usepackage{booktabs}
\usepackage{multirow} 
\usepackage{url}
\usepackage{xcolor}
\usepackage{colortbl}
\definecolor{mygray}{gray}{.9}
\newcommand\blfootnote[1]{%
  \begingroup
  \renewcommand\thefootnote{}\footnote{#1}%
  \addtocounter{footnote}{-1}%
  \endgroup
}
\AtBeginDocument{%
  }

\setcopyright{acmcopyright}
\copyrightyear{2023}
\acmYear{2023}
\acmDOI{XXXXXXX.XXXXXXX}

\acmSubmissionID{1615}

\RequirePackage[
  datamodel=acmdatamodel,
  style=acmnumeric,
  ]{biblatex}

\begin{document}

\title{ScribbleVC: Scribble-supervised Medical Image Segmentation with Vision-Class Embedding}

\author{Zihan Li}
\email{zhanli@uw.edu}
\affiliation{%
  \institution{Xiamen University, \\University of Washington}
  \city{Seattle}
  \country{USA}
  \postcode{98101}}

\author{Yuan Zheng}
\email{zhengyuan@stu.xmu.edu.cn}
\affiliation{%
  \institution{Xiamen University}
  \city{Xiamen}
  \country{China}
  \postcode{361005}}

\author{Xiangde Luo}
\email{xiangde.luo@std.uestc.edu.cn}
\affiliation{%
  \institution{University of Electronic Science and Technology of China}
  \city{Chengdu}
  \country{China}
  \postcode{611731}}

\author{Dandan Shan}
\email{shandd@stu.xmu.edu.cn}
\affiliation{%
  \institution{Xiamen University}
  \city{Xiamen}
  \country{China}
  \postcode{361005}}

\author{Qingqi Hong}
\email{hongqq@xmu.edu.cn}
\affiliation{%
  \institution{Xiamen University, COCHE}
  \city{Xiamen}
  \country{China}
  \postcode{361005}}

\renewcommand{\shortauthors}{Li, et al.}

\begin{abstract}
  Medical image segmentation plays a critical role in clinical decision-making, treatment planning, and disease monitoring. However, accurate segmentation of medical images is challenging due to several factors, such as the lack of high-quality annotation, imaging noise, and anatomical differences across patients. In addition, there is still a considerable gap in performance between the existing label-efficient methods and fully-supervised methods. To address the above challenges, we propose ScribbleVC, a novel framework for scribble-supervised medical image segmentation that leverages vision and class embeddings via the multimodal information enhancement mechanism. In addition, ScribbleVC uniformly utilizes the CNN features and Transformer features to achieve better visual feature extraction. The proposed method combines a scribble-based approach with a segmentation network and a class-embedding module to produce accurate segmentation masks. We evaluate ScribbleVC on three benchmark datasets and compare it with state-of-the-art methods. The experimental results demonstrate that our method outperforms existing approaches in terms of accuracy, robustness, and efficiency. The datasets and code are released on GitHub.\blfootnote{Corresponding author: Qingqi Hong. Li, Z and Zheng, Y have the equal contribution.}\footnote{\href{https://github.com/HUANGLIZI/ScribbleVC}{https://github.com/HUANGLIZI/ScribbleVC}}
\end{abstract}



\maketitle
\vspace{-4mm}
\section{Introduction}
Medical image segmentation plays a crucial role in medical image analysis, particularly in clinical practice where accurate segmentation is necessary for diagnosis and treatment planning. However, achieving accurate segmentation results for complex organs with intricate organizational structures remains a challenge, often requiring manual or semi-automatic methods. Recent studies have demonstrated the potential of deep learning for automatic medical image segmentation. However, creating high-quality medical image datasets is hampered by two issues: the high cost of expert annotation and the difficulty in obtaining high-quality medical images. These challenges limit the practical application of medical image segmentation models. To address these issues, researchers have started exploring label-efficient methods such as using scribble annotations for training. This approach shows promise in improving the performance of medical image segmentation models while reducing the need for expensive and time-consuming expert segmentation annotations and insufficient image annotations. Valvano et al. \cite{MAAG} proposed a scribble-supervised segmentation model based on multi-scale GAN and attention gates by introducing an unpaired segmentation mask, which requires additional annotation masks for model training. Meanwhile, Luo et al. \cite{luo2022scribble} proposed a scribble-supervised segmentation model by training a dual branch network and dynamically mixing pseudo-label supervision. In addition, Cyclemix \cite{Zhang_2022_CycleMix} is used to generate mixed images and regularization the model using circular consistency to perform medical image segmentation based on scribble supervision. While using scribble annotations for training can reduce the need for expensive and time-consuming expert segmentation annotations and insufficient image annotations, the imprecise nature of these labels can limit the accuracy of the resulting segmentation models. The limited supervised signals provided by scribble annotations can hinder the model's ability to learn the necessary visual features required for accurate medical image segmentation. Moreover, medical images often suffer from various quality defects that can adversely affect performance compared to fully supervised methods.

\begin{figure}[ht]
\setlength{\abovecaptionskip}{1mm}
  \centering
  \includegraphics[width=\linewidth]{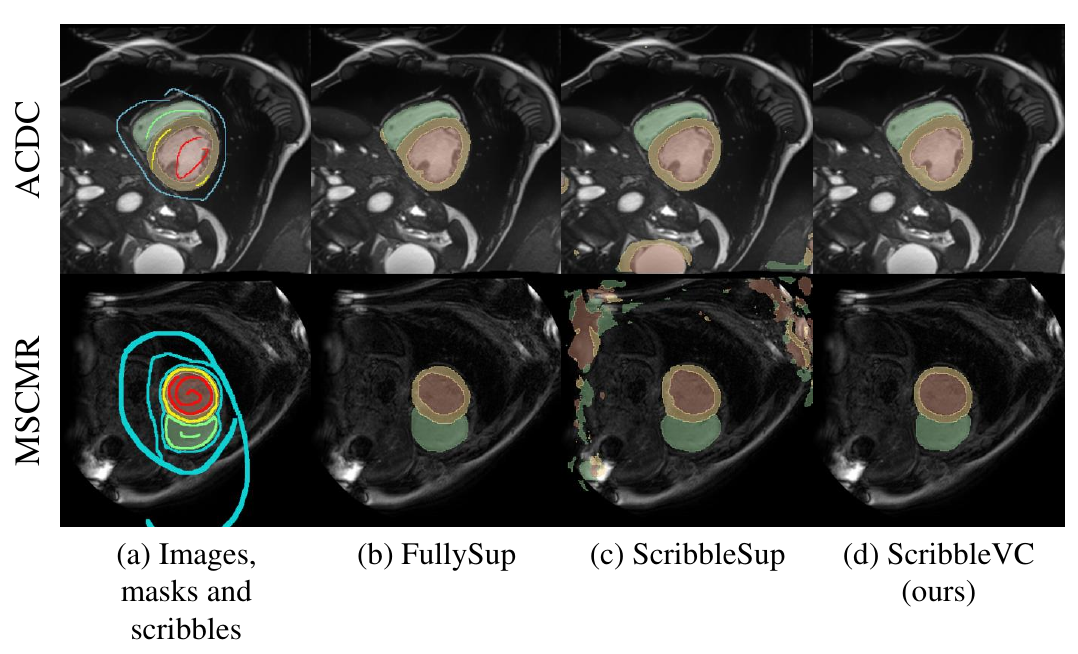}
  \caption{Performance comparison of segmentation results produced by different methods: (a) the input images, masks, and scribble annotations, (b) fully-supervised UNet++ \cite{unet++},  (c) scribble-supervised UNet++, and (d) ScribbleVC (ours).}
  \label{comparison}
  \vspace{-6mm}
\end{figure}
To effectively address these above issues, we propose the scribble-supervised model ScribbleVC that utilizes scribble-annotated images and visual class embedding features for training. To enhance the visual features of medical images, our model learns additional class embedding from category information. To address the issue of insufficient expert notes, we adopt scribble-supervised learning, which enables the model to extract features from scribble annotations while reducing the reliance on costly expert annotations. To better extract visual features, we design a CNN-Transformer encoder that unifies global and local features of the image. Our model incorporates two separate decoders, which extract CNN-style and Transformer-style features respectively, to fully utilize the information provided by scribble annotations. These decoders are supervised by the scribble annotations to ensure consistency between the two feature types. We incorporate class embedding features into our model to address the issue of low-quality medical images. Since category information is already present in the scribble annotations, our approach explicitly utilizes the categories, which helps segmentation even in the presence of quality defects. We obtain class embedding features through encoding rules rather than additional encoders, which reduces the number of parameters in the model. The multimodal information enhancement mechanism utilizes visual features and category information and improves pseudo labels through visual-class multimodal features. Overall, the main contributions of this paper are as follows:
 
\begin{itemize}
    \item We propose a brand new model (ScribleVC) for medical image segmentation with visual class embedding. To our knowledge, it is the first exploration of scribble-supervised models for visual-class embedding.
    \item We propose a multimodal information enhancement mechanism to introduce category feature information into visual features. In addition, we uniformly utilize CNN and Transformer features to achieve better visual feature extraction.
    \item To evaluate the performance of ScribbleVC, our study conducts experiments on the ACDC, MSCMRseg, and NCI-ISBI datasets. The results show that ScribbleVC has superior segmentation performance than other state-of-the-art methods, achieving a Dice score of 88.4\%, 86.8\%, and 79.8\% respectively.
\end{itemize}

\section{Related work}

\subsection{Medical Image Segmentation}
In the research and application of medical image segmentation technology\cite{hong2023distance}\cite{shan2023coarse}\cite{xu2022mrdff}, deep learning-based medical image segmentation technology is one of the current research hotspots. Deep learning algorithms are more adaptable to new pathological changes and different image qualities and can handle complex medical image segmentation tasks. Secondly, deep learning algorithms \cite{ronneberger2015u}\cite{milletari2016v}\cite{isensee2021nnu}\cite{luo2022semi} can process medical images that contain common problems such as noise, artifacts, and motion artifacts. Meanwhile, with the continuous development of computer technology, the real-time performance and accuracy of medical image segmentation algorithms based on deep learning have also been greatly improved.  
Typical network structures include U-Net\cite{ronneberger2015u}, etc. U-Net is a classic medical image segmentation model based on convolutional neural networks. The advantage of fully supervised medical image segmentation is that it can achieve high-precision segmentation results, especially in the case of a large amount of annotated data \cite{kirillov2023segment}\cite{qiu2023cor}. However, fully supervised methods typically require a large amount of annotated data for training, which is a bottleneck in the field of medical image segmentation \cite{wang2023SwinMM}. Due to the complexity and diversity of medical images, manually annotating data requires a significant amount of time and effort from professional doctors. In addition, annotated data may be very limited or difficult to obtain. How to train high-performance medical image segmentation models with as little annotated data as possible has also become an important challenge. Luo et al.\cite{luo2021urpc} utilized a pyramid prediction network and multi-scale uncertainty correction to learn from unlabeled data. 
\vspace{-4mm}
\subsection{Scribble-supervised Image Segmentation}
The weakly supervised learning method is another method to solve the problem of insufficient annotation data. The weakly supervised learning method is a training method using partially labeled data or weak supervised signals, which can improve the performance of the model in the case of limited labeled data. Weakly supervised learning methods can be divided into many types, such as tag noise-based methods, image-level annotation-based methods, and scribble-based methods. Scribble annotation refers to users manually drawing simple lines or scribbling to annotate the position of objects in an image. In the field of medical image segmentation, manual annotation data is usually provided in the form of points, lines, or regions. Ji et al. \cite{ji2019scribble} proposed a scribble-based hierarchical weakly supervised learning method for brain tumor segmentation, which combined weakly annotations for model training, including scribbles on the whole tumor and healthy brain tissue and the global labels for each Substructure. Valvano et al. \cite{MAAG} proposed a scribble-supervised segmentation model based on multi-scale GAN and attention gates by introducing an unpaired segmentation mask. These methods typically require additional dense annotations for model training. Therefore, we explored the impact of mask and scribble ratios on performance in our study.

\begin{figure*}[!h]
  \setlength{\abovecaptionskip}{-1mm}
  \centering
    \includegraphics[width=0.86\textwidth]{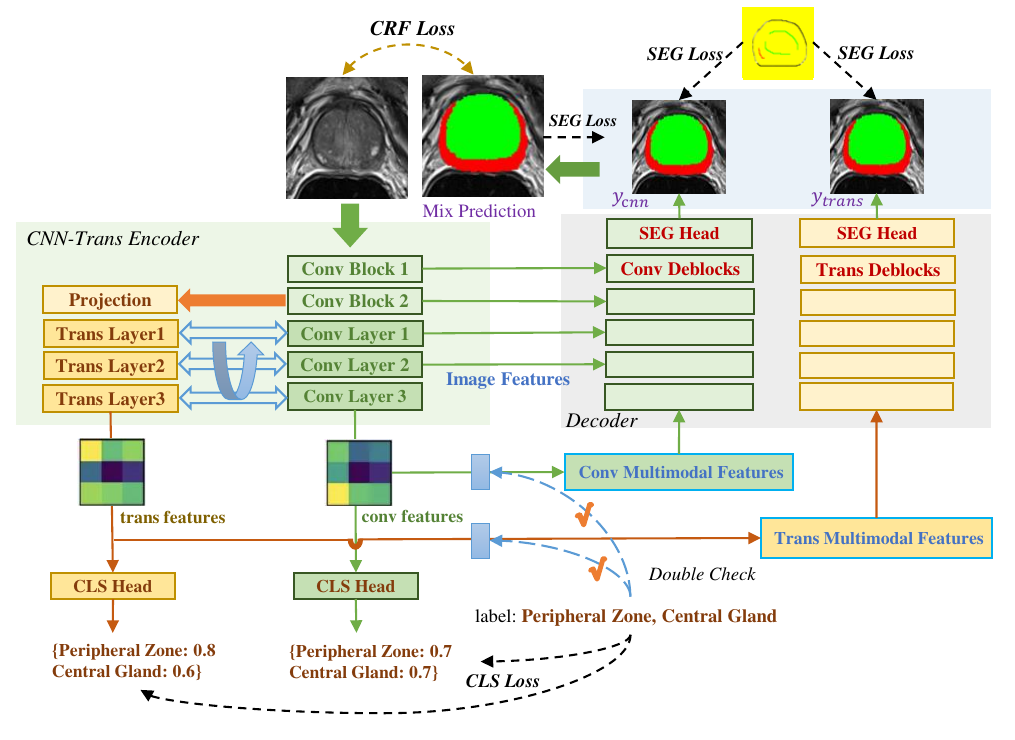}
  \caption{Overview of ScribbleVC, consisting of hybrid encoder-decoder and multimodal information enhancement module.}
  \label{fig:vlnetwork}
  \vspace{-4mm}
\end{figure*}
\vspace{-2mm}
\subsection{Multimodal learning}
Real-world information is often conveyed through multiple modalities, including images, videos, text, speech, and others \cite{Yang2023NeRCo}\cite{li2023chatdoctor}. Multimodal learning aims to identify the optimal feature representations from these diverse sources of information. In the area of natural image processing, the integration of images and other modal data has been widely used in semantic interpretation tasks such as Media Captioning 
\cite{wang2018video}\cite{lu2018neural}, Visual Question Answering 
\cite{hu2020iterative}\cite{marino2019ok}, Text-Image Retrieval \cite{wang2016learning} , and Text-to-Image Generation \cite{rombach2022high}\cite{rombach2022high}. In medical image analysis, different modalities often refer to imaging data acquired from various devices, such as positron emission tomography (PET), magnetic resonance imaging (MRI), and computed tomography (CT). Different modalities can represent complementary image features and information of the same object, and their synergistic cooperation can provide more comprehensive diagnostic information. Xue et al. \cite{xue2021multi} fed PET and CT images into a shared downsampling block to eliminate misleading features. Fu et al. \cite{fu2021multimodal} proposed a multimodal spatial attention module to emphasize the tumor-related regions in PET-CT images and suppress the irrelevant areas.
CLIP \cite{radford2021learning}, which predicts image categories by computing the similarity between images and texts, has gained widespread popularity among researchers. Therefore, recent works \cite{huang2021gloria}\cite{li2023lvit} in the field of medical image analysis have also attempted to incorporate textual information to improve performance in relevant tasks. For instance, GLoRIA \cite{huang2021gloria} learned global and local representations by comparing subregions of images and words in radiology reports. Zhou et al. \cite{zhou2022generalized} performed generalized radiograph representation learning by cross-supervising between medical images and radiology reports. However, different from the above methods, ScribbleVC does not require additional text annotations but can achieve multimodal interaction by extracting category information from the image.
\vspace{-4mm}
\section{Our approach}
Due to the limitations of annotation information in scribble annotations, we incorporate category feature information to assist in the extraction of image features. We design a Scribble-supervised model, ScribbleVC,  which is a multimodal model that consists of two main components: a hybrid encoding and decoding structure and a multimodal information enhancement module, as shown in Figure \ref{fig:vlnetwork}. 
Our proposed method utilizes a hybrid encoder that combines a convolutional neural network and Transformer to encode the input medical image. This encoder generates two types of feature representations: CNN image features and Transformer image features. To enhance the segmentation accuracy, our method incorporates known category information into these feature representations. It is achieved by extracting class embedding features and adding them to the corresponding features. This operation results in the formation of CNN multimodal features and Transformer multimodal features. Two separate decoders are designed to handle the differences in feature representations between the CNN and Transformer. To further improve visual information transmission, we introduce residual connections between the encoding and decoding modules of the CNN. Our method utilizes scribble supervision, pseudo-label supervision, and category supervision on different branches to generate a final segmentation result.
\begin{figure*}[!ht]
\setlength{\abovecaptionskip}{-1mm}
  \centering
  \includegraphics[width=0.86\textwidth]{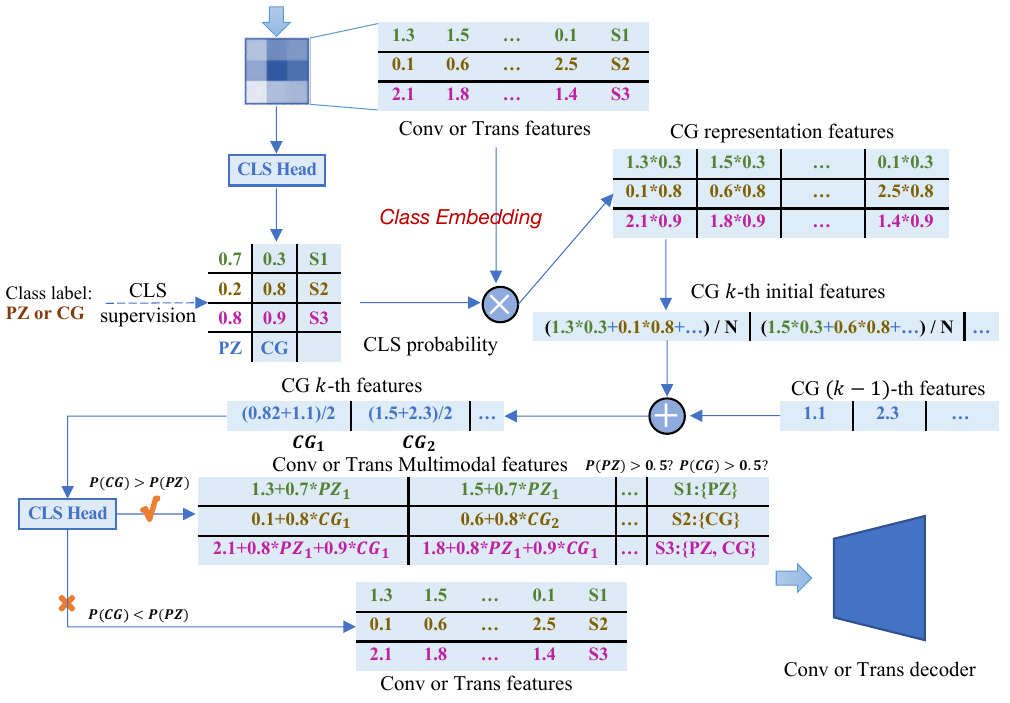}
  \caption{Overview of Multimodal Information Enhancement.}
  \vspace{-4mm}
\end{figure*}
\vspace{-2mm}
\subsection{Hybrid Encoder-Decoder}
\subsubsection{CNN-Trans Encoder}
Visual descriptors can be categorized into local features and global representations. Local features are compact vector representations of the image's local neighborhood, while global representations include shape descriptors, contour representations, and distant object types. In deep learning, convolutional neural networks leverage convolutional operations to construct multi-layer networks and collect local features, preserving them as feature maps. On the other hand, Transformer aggregates global representations by compressing patch embeddings via cascaded self-attention modules. To leverage the advantages of both local features and global representations, we propose a hybrid encoder that combines convolutional neural networks and Transformers. By exploiting the complementary nature of the two styles of features, the hybrid encoder inputs local context from the CNN branch into the feature map, enhancing the local perception ability of the Transformer branch. Similarly, the global features from the Transformer branch are gradually fed back to patch embedding, enriching the global representation of the CNN branch. This process also enables interaction between convolutional neural network feature information and Transformer feature information.
\begin{eqnarray} 
    &x_{trans,i+1} = Trans_{i}(x_{cnn,i}, x_{trans,i})
    \label{3-1.1}\\
    &x_{cnn,i+1} = Conv_{i}(x_{cnn,i}, x_{trans,i+1})
    \label{3-1.2}
\end{eqnarray}
where $x_{trans,i+1}$ denotes the output of the i-th Trans layer, whose inputs are $x_{cnn,i}$ and $x_{trans,i}$. And $x_{trans,i+1}$ will be the input of the i-th Conv layer. The output of the i-th Conv layer is $x_{cnn,i+1}$.

\subsubsection{Classification Head}
It is worth noting that to achieve the automatic generation of category feature information, two classification heads are designed at the tail of the encoder, which is respectively used to process CNN image features and Transformer image features. The classification head can automatically generate the category information contained in images, thereby achieving multimodal information enhancement.
\begin{eqnarray} 
    &y_{cls,trans} = CLSHead_{trans}(x_{trans,3})
    \label{3-1.1}\\
    &y_{cls,cnn} = CLSHead_{cnn}(x_{cnn,3})
    \label{3-1.2}
\end{eqnarray}
where $CLSHead_{trans}$ consists of one LayerNorm layer and one linear layer. And $CLSHead_{cnn}$ consists of one Conv2d layer and one AvgPool layer. The inputs of $CLSHead$ are $x_{trans,3}$ and $x_{cnn,3}$.

\subsubsection{Segmentation Decoder}
In the decoder section, we design the CNN decoder and Transformer decoder for processing different types of multimodal features. Both decoders utilize deconvolution to perform upsampling operations to ensure the reproducibility of model performance. The difference is that the encoder and decoder parts of the CNN have added residual connections to ensure that local features of the image can be better learned by the model. Due to the global nature of the category features imposed by the multimodal information enhancement mechanism, no additional residual connections were added to the Transformer encoder and decoder. Finally, the outputs of the CNN-branch decoder and the Transformer-branch decoder form a mixed prediction result, which is used as a pseudo label to realize the weakly supervised learning.
\vspace{-2mm}
\subsection{Multimodal Information Enhancement}
To fully utilize the category information in the Scribble labels, we propose a multimodal information enhancement mechanism. First, we extract the feature vectors of the category. Next, the hybrid encoder undergoes feature interaction and outputs image feature vectors with both global and local information. These image feature vectors are then predicted by the classification head to obtain prediction probabilities. We then use category embedding to multiply the predicted probability of each category by the image feature vector, resulting in the characteristic features of each sample in the batch corresponding to the category. Finally, we calculate the mean of the characteristic features of the batch to obtain the category feature vector of the corresponding category in the $k$-th batch.
To update the historical category feature vector, the second step involves averaging the category feature vector with the corresponding vector from the previous batch, followed by prediction using the classification header. If the new category feature vector outperforms the previous vector in the prediction results, it is updated as the new historical category feature vector; otherwise, the previous vector remains unchanged.
In the third step, we combine class embeddings with image feature vectors to obtain multimodal fusion feature vectors. For a sample, if the predicted class probability of its image feature vector is greater than 0.5, we consider its expected prediction value for that class to be 1, indicating that the sample can introduce class feature vectors. If all the predicted expected values of the categories for the sample are 1, and they meet the conditions described in the second step, we use the prediction probability of the sample for the category as the weight of the category feature vector. After the weighted sum, we add the image feature vector to obtain the mixed feature. However, if a category with a predicted expected value of 1 for the sample does not meet the conditions described in the second step, the fusion feature vector will not be updated, and the feature vector output by the encoder will still be used as input to the decoder. The design aims to highlight all categories together when enhancing image feature vectors, as adding only one may lead to imbalanced category features.

During training, the model retains historical category feature vectors, which are then used in the testing phase to replace the category feature vectors. In the testing phase, the model does not extract category feature vectors from each image feature vector in the test set, and it does not perform the second step of detecting category feature vectors and updating historical category feature vectors. Instead, it only calculates the category prediction probability for each image feature vector in the test set.
To generate predictions, the model multiplies the historical category feature vector corresponding to the predicted category by the predicted probability value of the image feature vector in that category. It then weights and sums all historical category feature vectors that meet the conditions, and adds the resulting vector to the image feature vector as input to the decoder.
\vspace{-2mm}
\subsection{Training Strategy and Loss Function}
The overall training strategy is divided into four parts. The first part is the supervision of scribble annotations, the second part is the supervision of unlabeled pixels by using the threshold-based mechanism, the third part is the loss of gating condition random field, and the fourth part is the classification supervision of category.
\begin{eqnarray}
&L_{ss}\left(s, y_{c n n}, y_{trans}\right)=\frac{L_{c e}\left(y_{c n n}, s\right)+ L_{c e}\left(y_{trans}, s\right)}{2}
\label{eq:Lss}
\end{eqnarray}
In the first part, the segmentation results of the convolutional neural network and Transformer are supervised by partial cross entropy function $L_{ss}$ with scribble annotation. Among them, $y_{cnn}$is the prediction result of the convolutional neural network branch, and $y_{trans}$is the predicted result of the Transformer branch. $L_{ce}$is a partial cross entropy function, which is defined as:
\begin{eqnarray}
L_{ce}(y, s) & = & \sum_{i \in \Omega_{l}} \sum_{k \in K}-s_{i}^{k} \log \left(y_{i}^{k}\right)
\label{eq:Lce}
\end{eqnarray}
where $K$ is the set of categories in the image, and $Omega_{l}$is the set of labeled pixels in the scribble $s$; $s_i^k$ and $y_i^k$ are the probability that the $i$-th pixel in the scribble belongs to the $k$-th class, and the probability that the $i$-th pixel in the prediction results belongs to the $k$-th class, respectively.

In the second part, we employ a threshold-based pseudo-labeling mechanism to supervise unlabeled pixels. The dual-branch network is utilized to generate two sets of predictions with different attentional focuses, namely local information and center position offset. The mixed prediction is used to supervise both branches. The pseudo label $Y$ is generated using a threshold-based approach to reduce errors. It is achieved by combining the predicted probabilities from both branches according to the following formula:
\begin{eqnarray}
Y = \alpha \times (y_{cnn}>t) \times y_{cnn}+(1-\alpha) \times (y_{trans}>t) \times y_{trans}
\end{eqnarray}
To include dynamic prediction results for a given pixel, the prediction probabilities of both branches at that pixel must exceed the threshold $t$. This criterion ensures that unreliable predicted pixels are excluded from the dynamic prediction results. In this study, the threshold was set to 0.5. Additionally, a random number $\alpha$ is generated for each batch, with a range of (0,1). This strategy allows the convolutional neural network branches and Transformer branches to learn from each other through pseudo-labels, and dynamic mixing improves the diversity of the pseudo-labels. The threshold setting helps prevent prediction errors from misleading the model through the use of pseudo-labels.

To balance the local and global information provided by CNN and Transformers, we proposes a strategy to limit the gradient flow between the two branches while avoiding consistency learning. It maintains the independence of the two branches and allows the supervised signal to propagate to all unlabeled pixels. The mixed prediction results are then used to supervise both branches during training. The dynamic prediction $Y_t$ supervision approach amplifies the supervised signal from limited annotated pixels to the entire image. The formula for dynamic prediction supervision is:
\begin{eqnarray}
L_{pl} = \frac{L_{dice}\left(y_{cnn}, \mathrm{argmax}(Y_t)\right) + L_{dice}\left(y_{trans}, \mathrm{argmax}(Y_t)\right)}{2}
\end{eqnarray}
The third part introduces the gated conditional random field loss. Gated conditional random field loss is commonly used in the training of weakly supervised image segmentation methods. It helps to eliminate the influence of irrelevant pixels on the classification of the current pixel. Furthermore, it places more emphasis on the semantic boundary rather than the semantic relationship between regions. This simplifies the process of combining a conditional Random field and CNN. Additionally, it does rely on high-dimensional filters. The gated conditional random field loss is defined as:
\begin{eqnarray}
L_{crf} = \sum_{i = 1}^{N} \sum_{j = 1}^{N} w_{ij} \cdot \phi(x_i, x_j) \cdot (Y_i - Y_j)^2
\end{eqnarray}
where $N$ is the number of pixels, and $w_{ij} $ is the gating function to mask unexpected pixel positions.  The similarity between the pixels $x_i$ and $x_j$ is measured by the function $\phi(x_i, x_j)$. Additionally, $Y_i$ and $Y_j$ are the prediction probability value of pixels $i$ and $j$, respectively.

The fourth part improves the accuracy of category features by applying a classification loss to the encoded features. The category loss is defined as:
\begin{eqnarray}
    L_{cls} = \mathrm{avg}\left(L_{bce}\left(p_{cnn}, c\right), L_{bce}\left(p_{trans}, c\right)\right)
\end{eqnarray}
where $p_{cnn}$, and $p_{trans}$ represent the prediction probability of convolutional neural networks and Transformer networks, respectively, while $c$ represents the actual category of input images. Because an input image may correspond to multiple categories, the predicted probability and actual category adopt a binary loss of $L_{bce}$:
\begin{eqnarray}
    L_{bce} = -\sum_{i = 1}^{N}[c_{i} \ln (p_{i})+(1-c_{i}) \ln (1-p_{i})]
\end{eqnarray}
where $N$ represents the total number of samples. To ensure probability value can predict multiple categories simultaneously, the prediction probability $p$ is obtained through the sigmoid function instead of the softmax function.
Finally, the total loss function is:
\begin{eqnarray}
L_{total} & = & \lambda_{1} \times L_{ss}+\lambda_{2} \times L_{pl}+\lambda_{3} \times L_{crf}+\lambda_{4} \times L_{cls}
\label{eq:text_total}
\end{eqnarray}
where $\lambda_{1-4}$ are the weights of each part of the loss to balance different supervised losses.

\section{Experiments}
\begin{figure*}[htbp]
\setlength{\abovecaptionskip}{0mm}
\centering 
\includegraphics[width=0.92\textwidth]{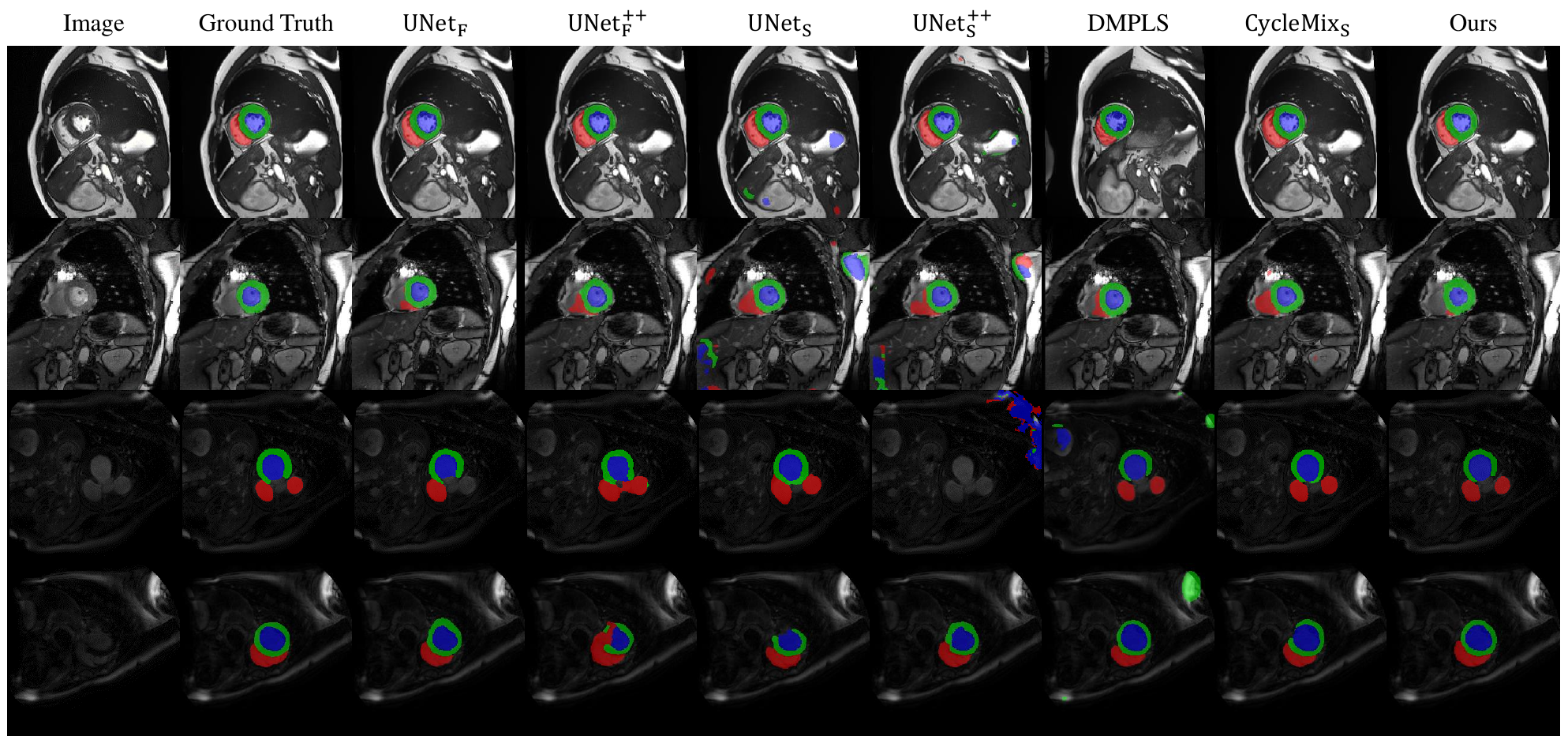}
\caption{Qualitative comparison between our method (ScribbleVC) and other state-of-the-art methods on ACDC and MSCMRseg datasets. Subscripts \textit{F} and \textit{S} indicate segmentation models are trained with dense annotations or scribble annotations.}
\label{results}
\vspace{-5mm}
\end{figure*}
\subsection{Setup}
\subsubsection{Datasets}
\textbf{ACDC dataset} \cite{acdc} includes 100 cine-MRI scans with manual scribble annotations for RV, LV, and MYO supplied by \cite{MAAG}. The scans are divided into sets of 70, 15, and 15 for training, validation, and testing. We split the training set into two halves, 35 images with scribble labels and 35 masks with full annotations. It is worth noting that the corresponding masks are not used in training. \textbf{MSCMRseg dataset} \cite{mscmr1, mscmr2} includes Late Gadolinium Enhancement MRI scans from 45 cardiomyopathy patients, each with scribble annotations of LV, MYO, and RV provided by \cite{Zhang_2022_CycleMix}. The 45 scans are randomly partitioned into three sets: 25 for training, 5 for validation, and 15 for testing. \textbf{NCI-ISBI dataset} \cite{Clark2013Cancer} is from ISBI 2013 Prostate Magnetic Resonance Imaging Challenge. There are 80 volumes in the NCI-ISBI dataset, which are divided into a training set and a test set of 3:1. All the labels in the training set are scribble annotations, and the category information is provided by the scribble labels. Category information is only used as a supervisory signal during training and is not provided during testing.

\subsubsection{Implementation details}
The model was implemented using Pytorch and trained on one NVIDIA RTX 3090. To expand the training set, we applied random rotation, flipping, and noise to the images. The learning rate is fixed at 1e-4, and the weight decay is set to 0.0005. Our model is trained with AdamW optimizer for 300 epochs in the experiments. We empirically set the weights $(\lambda_1, \lambda_2, \lambda_3, \lambda_3)$ to (1, 0.5, 0.1, 0.1) in Eqn. \ref{eq:text_total}. For all datasets, the Dice coefficient (Dice) is used as an evaluation metric.
\vspace{-2mm}
\begin{table}[!ht]\footnotesize
\setlength{\abovecaptionskip}{0mm}
  \caption{Performance comparison between our method (ScribbleVC) and other state-of-the-art methods on ACDC. Bold denotes the best performance.}
    \resizebox{0.98\columnwidth}{!}{%
    \begin{tabular}{l|c|ccc|c}
    \toprule[1pt]
    Methods & Data  & LV & MYO & RV & Avg \\ \hline
    \multicolumn{6}{l}{\textbf{35 scribbles}} \\ \hline
    UNETR\cite{Hatamizadeh_2022_WACV} & scribbles & .688 & .330 & .180 & .399 \\
    SwinUNETR\cite{swinunetr} & scribbles & .768 & .683 & .640 & .697 \\
    SwinUNet\cite{swinunet} & scribbles & .862 & .768 & .735 & .788 \\ 
    TransUNet\cite{chen2021transunet} & scribbles & .599 & .475 & .428 & .501 \\ 
    TFCNs\cite{li2022tfcns} & scribbles & .703 & .614 & .619 & .645 \\ 
    UNet$_{pce}$\cite{lin2016scribblesup} & scribbles & .624 & .537 & .526 & .562 \\
    UNet$_{wpce}$\cite{MAAG} & scribbles & .784 & .675 & .563 & .674 \\
    UNet$_{ustr}$\cite{liu2022weakly} & scribbles & .605 & .599 & .655 & .620 \\
    UNet$_{mloss}$\cite{kim2019mumford} & scribbles & .873 & .812 & .833 & .839 \\
    UNet$_{em}$\cite{EM} & scribbles & .789 & .761 & .788 & .779 \\
    UNet$_{crf}$\cite{crf} & scribbles & .766 & .661 & .590 & .672 \\
    UNet$^{+}_{pce}$\cite{unet+} & scribbles & .785 & .725 & .746 & .752 \\
    UNet$^{++}_{pce}$\cite{unet++} & scribbles & .846 & .787 & .652 & .761 \\
    MixUp\cite{mixup} & scribbles & .803 & .753 & .767 & .774 \\
    Cutout\cite{cutout} & scribbles & .832 & .754 & .812 & .800 \\
    CutMix\cite{cutmix} & scribbles & .641 & .734 & .740 & .705 \\
    Puzzle Mix\cite{puzzlemix} & scribbles & .663 & .650 & .559 & .624 \\
    Co-mixup\cite{comixup} & scribbles & .622 & .621 & .702 & .648 \\
    CycleMix$_S$\cite{Zhang_2022_CycleMix} & scribbles & .883 & .798 & .863 & .848 \\
    \rowcolor{mygray}
    \textbf{ScribbleVC} & scribbles & \textbf{.914} & \textbf{.866} & \textbf{.870} & \textbf{.884} \\ \hline
    \multicolumn{6}{l}{\textbf{35 scribbles + 35 unpaired masks}} \\ \hline
    UNet$_D$\cite{MAAG} & mixed & .404 & .597 & .753 & .585 \\
    PostDAE\cite{postdae} & mixed & .806 & .667 & .556 & .676 \\
    ACCL\cite{accl}  & mixed & .878 & .797 & .735 & .803 \\
    MAAG\cite{MAAG}  & mixed & .879 & .817 & .752 & .816 \\ \hline
    \multicolumn{6}{l}{\textbf{35 masks}} \\ \hline
    UNet$_F$\cite{unet} & masks & .892 & .830 & .789 & .837 \\
    UNet$^+_F$\cite{unet+} & masks & .849 & .792 & .817 & .820 \\
    UNet$^{++}_F$\cite{unet++} & masks & .875 & .798 & .771 & .815 \\
    Puzzle Mix$_F$\cite{puzzlemix} & masks &.849 & .807 & .865 & .840 \\
    VT-UNet & masks & .895 & .807 & .804 & .836 \\
    UNETR\cite{Hatamizadeh_2022_WACV} & masks & .926 & .844 & .845 & .872 \\
    SwinUNet\cite{swinunet} & masks & .900 & .812 & .818 & .843 \\
    \bottomrule[1pt]
    \end{tabular}%
    }
  \label{ACDCresult}%
  \vspace{-6mm}
\end{table}
\vspace{-4mm}
\subsection{Performance Comparison with Other State-Of-The-Art Methods}
To demonstrate the comprehensive segmentation performance of our method, we compare ScribleVC with different SOTA methods.

1) Transformer-based fully-supervised segmentation methods, including UNEt TRansformers (UNETR) \cite{Hatamizadeh_2022_WACV}, Swin UNEt TRansformers (SwinUNETR) \cite{swinunetr}, SwinUNet \cite{swinunet}, TransUNet \cite{chen2021transunet} and TFCNs \cite{li2022tfcns} which are the medical image segmentation models utilizing a combination of convolutional layers and Transformers.

2) Different scribble-supervised strategies on UNet: partial cross-entropy loss (pce) \cite{lin2016scribblesup}, 
weighted partial cross-entropy loss (wpce) \cite{MAAG}, uncertainty self-ensembling and transformation-consistent regularization (ustr) \cite{liu2022weakly}, mumford–shah Loss (mloss) \cite{kim2019mumford}, 
entropy minimization (em) \cite{EM}, and conditional random field (crf) \cite{crf}.

3) Different scribble-supervised frameworks with the same loss: the partial cross-entropy loss on different variants of UNet$_{pce}$ \cite{lin2016scribblesup}, including UNet$^{+}_{pce}$ \cite{unet+} 
and UNet$^{++}_{pce}$ \cite{unet++}.

4) Different data augmentation: MixUp \cite{mixup}, Cutout \cite{cutout}, CutMix \cite{cutmix}, Puzzle Mix \cite{puzzlemix}, Co-mixup \cite{comixup}, CycleMix$_S$ \cite{Zhang_2022_CycleMix}.
Second, we also compare some adversarial learning methods, including UNet$_D$ \cite{MAAG}, PostDAE \cite{postdae}, ACCL \cite{accl}, and MAAG \cite{MAAG}.

Finally, we investigate the fully-supervised methods: UNet$_F$ \cite{unet}, UNet$^+_F$ \cite{unet+},  UNet$^{++}_F$ \cite{unet++}, Puzzle Mix$_F$ \cite{puzzlemix} and CycleMix$_F$ \cite{Zhang_2022_CycleMix}. 
\vspace{-4mm}
\begin{table}[ht!]\footnotesize
\setlength{\abovecaptionskip}{0mm}
  \centering
  \caption{Performance comparison between our method (ScribbleVC) and other state-of-the-art methods on MSCMRseg. Bold denotes the best performance.}
  \resizebox{0.98\columnwidth}{!}{%
    \begin{tabular}{l|c|ccc|c}
    \toprule[1pt]
    Methods & Data  & LV & MYO & RV & Avg \\ \hline
    \multicolumn{6}{l}{\textbf{25 scribbles}} \\  \hline
    UNet$^+_{pce}$\cite{unet+} & scribbles & .494 & .583 & .057 & .378 \\
    UNet$^{++}_{pce}$\cite{unet++} & scribbles & .497 & .506 & .472 & .492 \\
    MixUp\cite{mixup} & scribbles & .610 & .463 & .378 & .484 \\
    Cutout\cite{cutout} & scribbles & .459 & .641 & .697 & .599 \\
    CutMix\cite{cutmix} & scribbles & .578 & .622 & .761 & .654 \\
    Puzzle Mix\cite{puzzlemix} & scribbles & .061 & .634 & .028 & .241 \\
    Co-mixup\cite{comixup} & scribbles & .356 & .343 & .053 & .251 \\
    DMPLS\cite{luo2022scribble} & scribbles & .881 & .644 & \textbf{.863} & .796 \\
    CycleMix$_S$\cite{Zhang_2022_CycleMix} & scribbles & .870 & .739 & .791 & .800 \\
    \rowcolor{mygray}
    \textbf{ScribbleVC} & scribbles & \textbf{.921} & \textbf{.830} & .852 & \textbf{.868} \\  \hline
    \multicolumn{6}{l}{\textbf{25 masks}} \\  \hline
    UNet$_F$\cite{unet} & masks & .850 & .721 & .738 &  .770 \\
    UNet$^+_F$\cite{unet+} & masks & .857 & .720 & .689 & .755 \\
    UNet$^{++}_F$\cite{unet++} & masks & .866 & .745 & .731 & .774 \\
    Puzzle Mix$_F$\cite{puzzlemix} & masks & .867 & .742 & .759 & .789 \\
    CycleMix$_F$\cite{Zhang_2022_CycleMix} & masks & .864 & .785 & .781 & .810 \\ \bottomrule[1pt]
    \end{tabular}%
    }
  \label{MSCMRresult}%
  \vspace{-3mm}
\end{table}%

As shown in Table \ref{ACDCresult} and Table \ref{MSCMRresult}, our ScribbleVC model outperforms a number of training strategies, model architectures, and data augmentation techniques based on UNet in scribble supervision. In particular, it outperforms the SOTA method CycleMix by a margin of 3.6\% (88.4\% vs 84.8\%) and 6.8\% (86.8\% vs 80.0\%) on ACDC and MSCMRseg, respectively, which demonstrates the effectiveness of incorporating Transformer global context to CNN local features in scribble-supervised semantic segmentation. Meanwhile, we found that the full-annotation-designed Transformer-based medical image segmentation models only achieved average performance on scribble data. In contrast, our ScribbleVC model can achieve superior performance by jointly leveraging local detailed information and global context. The ACDC results in the second section (scribbles + unpaired masks) of Table~\ref{ACDCresult} demonstrate significant performance improvements of ScribbleVC compared to other weakly-supervised methods. It can be observed that the Dice scores of all three categories of LV, MYO, and RV achieved by ScribbleVC have exceeded the previous best method (MAAG \cite{MAAG}). We believe those methods with additional unpaired masks could only learn limited shape priors due to the vague of segmentation boundaries.

On the other hand, ScribbleVC can overcome this limitation by utilizing the self-attention mechanism of Transformers to learn global shapes without additional fully-annotated masks. 
In the last section of Table~\ref{ACDCresult} and Table~\ref{MSCMRresult}, we also compared the proposed ScribbleVC with several fully-supervised learning methods on ACDC and MSCMR, including CycleMix with fully-supervised learning. As shown in the tables, the results with fully-supervised learning are better than those with scribble annotations plus additional unpaired masks because of the acquisition of pixel-wise relationships.
However, our ScribbleVC outperforms the fully-supervised methods at a lower annotation cost, demonstrating the great potential in medical image segmentation.
\vspace{-4mm}
\subsection{Comparison with Pseudo-label Generating Methods on the ACDC dataset}

To compare our ScribbleVC with other pseudo-label generating methods, we employed UNet with only partial cross-entropy loss (pce) \cite{lin2016scribblesup} as the base segmentation network architecture plus: 1) using pseudo label generated by Random Walker (rw) \cite{grady2006random}, 2) incorporating pseudo-labeling plus label filtering named Scribble2Label (s2l) \cite{S2L}, 3) with dual-branch using dynamically mixed pseudo labels supervision (DMPLS) \cite{luo2022scribble}.
Additionally,  we also compare with TS-UNet \cite{can2018learning}, a variant of UNet+ with a combination of the random walker, dense CRF, and uncertain estimation.
\vspace{-2mm}
\begin{table}[!htbp]\footnotesize
\setlength{\abovecaptionskip}{0mm}
  \centering
  \caption{Comparison with pseudo-label generating methods on the ACDC dataset.}
  \resizebox{0.98\columnwidth}{!}{%
    \begin{tabular}{l|c|ccc|c}
    \toprule[1pt]
    Methods & Data  & LV & MYO & RV & Avg \\
    \midrule
    TS-UNet\cite{can2018learning} & scribbles & .479 & .408 & .272 & .386 \\
    UNet$_{pce}$\cite{lin2016scribblesup} & scribbles & .624 & .537 & .526 & .562 \\
    UNet$_{rw}$\cite{grady2006random} & scribbles & .840 & .730 & .802 & .791 \\
    UNet$_{s2l}$\cite{S2L} & scribbles & .767 & .715 & .765 & .820 \\
    DMPLS\cite{luo2022scribble} & scribbles & .875 & \textbf{.903} & .852 & .870 \\
    \rowcolor{mygray}
    \textbf{ScribbleVC} & scribbles & \textbf{.914} & .866 & \textbf{.870} & \textbf{.884} \\
    \bottomrule[1pt]
    \end{tabular}%
    }
  \label{pseudo}%
  \vspace{-2mm}
\end{table}%
\vspace{-2mm}

As shown in Table~\ref{pseudo}, some pseudo-label-based methods with scribble annotations can achieve reasonably good performance, with both S2L and DMPLS achieving 80\% or higher. Nevertheless, our method outperforms these methods by a significant margin, confirming the enhancement of pseudo-label generating of the CNN-Transformer synergy in our network.
\vspace{-2mm}
\subsection{Comparison with Scribble-supervised Methods on the NCI-ISBI dataset}
In this section, we compared our method with scribble-supervised segmentation methods in NCI-ISBI scribble-annotated medical images. Specifically, we employed UNet as the base segmentation network architecture with
partial cross-entropy loss (Scribblesup) \cite{lin2016scribblesup}, 
utilizing uncertainty-aware self-ensembling and transformation-consistent regularization (USTM) \cite{liu2022weakly}, 
using entropy minimization (SSEM) regularization \cite{EM}, incorporating pseudo-labeling plus label filtering named Scribble2Label (S2L) \cite{S2L}, 3D-UNet \cite{cciccek20163d}, SegNet \cite{badrinarayanan2017segnet} and CRF-RNN\cite{monteiro2018conditional}.
All baseline models are trained only on the labeled pixels of the scribble data. 
The results are reported in Table~\ref{tab:ISBIresult}. We found that 
our ScribbleVC model can achieve superior performance to other scribble-supervised and even fully-supervised methods by jointly leveraging local detailed information and global context, which demonstrates the effectiveness of incorporating Transformer global context to CNN local features in scribble-supervised semantic segmentation.
\vspace{-2mm}
\begin{table}[!ht]\tiny
\setlength{\abovecaptionskip}{0mm}
    \centering
    \caption{Comparison with scribble-supervised methods on the Prostate (NCI-ISBI)  dataset.} \label{tab:ISBIresult}
    \resizebox{0.98\columnwidth}{!}{%
    \begin{tabular}{l|c|cc|c}
    \toprule
        Methods & Data  & PZ & CG & Avg  \\ 
        \midrule
        Scribblesup\cite{lin2016scribblesup} & scribbles & .271 & .369 & .320  \\ 
        USTM\cite{liu2022weakly} & scribbles & .401 & .209 & .305 \\ 
        SSEM\cite{EM} & scribbles & .501 & .393 & .447 \\
        S2L\cite{S2L} & scribbles & .674 & .650 & .662 \\
        3D-UNet\cite{cciccek20163d} & scribbles &	.670 & .829 &.750\\
        SegNet\cite{badrinarayanan2017segnet}& scribbles &.720 & .837 &.778\\
        CRF-RNN\cite{monteiro2018conditional} & scribbles & .698 & \textbf{.863} &.781\\
        \rowcolor{mygray}
        \textbf{ScribbleVC} & scribbles & \textbf{.743} & .854 & \textbf{.798} \\ \midrule
        UNet$_F$ & masks & .723 & .832 & .778  \\
    \bottomrule
    \end{tabular}
    }
    \vspace{-4mm}
\end{table}
\vspace{-2mm}
\subsection{Ablation Experiments}
The section studies the effectiveness of different components of the proposed ScribbleVC, including CNN, Transformer, and CLS modules. Table~\ref{tab:ablation} reports the results.
Compared with \#1 with only convolutional neural network branches and \#2 with only Transformer branches, \#3 with both convolutional neural network and Transformer branches has better performance, indicating that the synergistic effect of convolutional neural network and Transformer has a promoting effect on the model. Compared to \#3, \#4 with multimodal information enhancement mechanism exhibits better performance, confirming the effectiveness of this mechanism.
\vspace{-2mm}
\begin{table}[ht!]\footnotesize
\setlength{\abovecaptionskip}{0mm}
  \centering
  \caption{Ablation study: ScribbleVC for image segmentation with different settings, including the CNN branch, the Transformer branch, and CLS module.} \label{tab:ablation}
   \resizebox{0.98\columnwidth}{!}{%
    \begin{tabular}{c|ccc|cc|c}
    \toprule[1pt]
   Models & CNN & Transformer & CLS & PZ & CG & Avg \\
   \midrule
    \#1  & \checkmark & $\times$   & $\times$   & .666 & .167 & .416 \\
    \#2  & $\times$   & \checkmark & $\times$   & .433 & .633 & .533 \\
    \#3  & \checkmark & \checkmark & $\times$   & .708 & .843 & .775 \\
    \#4  & \checkmark & \checkmark & \checkmark & \textbf{.743} & \textbf{.854} & \textbf{.798} \\
    \bottomrule[1pt]
  \end{tabular}}
  \vspace{-2mm}
\end{table}
\vspace{-4mm}
\subsection{Data Sensitivity Experiments}
The data sensitivity study investigates the performance of ScribbleVC with different numbers of scribble annotations during training. 
As shown in Table~\ref{sensitivity}, the performance of ScribbleVC has been boosted gradually as the number of scribble-annotated samples increases. Even with just 20 training samples with scribbles, our model can reach 75.1\%, which confirms that ScribbleVC is able to achieve satisfactory segmentation results with a relatively small amount of scribble annotations. The overall performance of ScribbleVC stabilized when the number of scribble annotations reached 40 (67\% of 60 scribbles). The best performance can be achieved by using all 60 scribble annotations, resulting in an accuracy of 79.8\%. 
\vspace{-6mm}
\begin{table}[!ht]\tiny
\setlength{\abovecaptionskip}{0mm}
  \centering
  \caption{Data sensitivity study: the performance of ScribbleVC with the different numbers of scribbles for training.}
  \resizebox{0.98\columnwidth}{!}{%
  {%
    \begin{tabular}{c|c|cc|c}
    \toprule
    Method & Scribble Data & PZ & CG & Avg \\
    \midrule
    ScribbleVC & 20 scribbles & .668 & .833 & .751 \\
    ScribbleVC & 30 scribbles & .713 & .834 & .773 \\ 
    ScribbleVC & 40 scribbles & .728 & .846 & .787 \\
    ScribbleVC & 50 scribbles & .726 & .859 & .792 \\
    ScribbleVC & 60 scribbles & \textbf{.743} & \textbf{.854} & \textbf{.798} \\
    \bottomrule
    \end{tabular}%
    }}
  \label{sensitivity}%
  \vspace{-2mm}
\end{table}%
\vspace{-4mm}
\section{Conclusion}
In this paper, we present ScribleVC, a novel model for medical image segmentation using scribble supervision. By leveraging category information from scribble labels, ScribleVC enhances the effectiveness of this annotation method. Our approach employs a multimodal information enhancement mechanism to incorporate category feature information into visual features. Additionally, we achieve improved visual feature extraction by leveraging both CNN and Transformer features. As the first exploration of scribble-supervised models for visual-class embedding, ScribleVC is a simple yet effective model that delivers high-quality pixel-level segmentation results. Experimental results show that our ScribleVC outperforms state-of-the-art methods on the ACDC, MSCMRseg, and NCI-ISBI datasets.\\
\noindent\textbf{Acknowledgments}: This work was supported in part by the Natural Science Foundation of Fujian Province of China (No. 2020J01006), the Open Project Program of State Key Laboratory of Virtual Reality Technology and Systems, Beihang University (No. VRLAB2022AC04), and ITC-InnoHK Projects at Hong Kong Centre for Cerebro Cardiovascular Health Engineering (COCHE).

\printbibliography

@String{Computing = "Computing" }

@String{Computer = "{IEEE} Computer" }

@String{Springer = "Springer-Verlag" }

@inproceedings{ji2019scribble,
  title={Scribble-based hierarchical weakly supervised learning for brain tumor segmentation},
  author={Ji, Zhanghexuan and Shen, Yan and Ma, Chunwei and Gao, Mingchen},
  booktitle={International Conference on Medical Image Computing and Computer-Assisted Intervention},
  pages={175--183},
  year={2019},
  organization={Springer}
}

@article{MAAG,
  title={Learning to segment from scribbles using multi-scale adversarial attention gates},
  author={Valvano, Gabriele and Leo, Andrea and Tsaftaris, Sotirios A},
  journal={IEEE Transactions on Medical Imaging},
  volume={40},
  number={8},
  pages={1990--2001},
  year={2021},
  publisher={IEEE}
}

@article{luo2022scribble,
  title={Scribble-Supervised Medical Image Segmentation via Dual-Branch Network and Dynamically Mixed Pseudo Labels Supervision},
  author={Luo, Xiangde and Hu, Minhao and Liao, Wenjun and Zhai, Shuwei and Song, Tao and Wang, Guotai and Zhang, Shaoting},
  journal={arXiv preprint arXiv:2203.02106},
  year={2022}
}

@InProceedings{Zhang_2022_CycleMix,
    author    = {Zhang, Ke and Zhuang, Xiahai},
    title     = {CycleMix: A Holistic Strategy for Medical Image Segmentation From Scribble Supervision},
    booktitle = {Proceedings of the IEEE/CVF Conference on Computer Vision and Pattern Recognition (CVPR)},
    month     = {June},
    year      = {2022},
    pages     = {11656-11665}
}

@article{transformer,
  title={Attention is all you need},
  author={Vaswani, Ashish and Shazeer, Noam and Parmar, Niki and Uszkoreit, Jakob and Jones, Llion and Gomez, Aidan N and Kaiser, {\L}ukasz and Polosukhin, Illia},
  journal={Advances in neural information processing systems},
  volume={30},
  year={2017}
}

@incollection{unet++,
  title={Unet++: A nested u-net architecture for medical image segmentation},
  author={Zhou, Zongwei and Rahman Siddiquee, Md Mahfuzur and Tajbakhsh, Nima and Liang, Jianming},
  booktitle={Deep learning in medical image analysis and multimodal learning for clinical decision support},
  pages={3--11},
  year={2018},
  publisher={Springer}
}

@inproceedings{crf,
  title={Conditional random fields as recurrent neural networks},
  author={Zheng, Shuai and Jayasumana, Sadeep and Romera-Paredes, Bernardino and Vineet, Vibhav and Su, Zhizhong and Du, Dalong and Huang, Chang and Torr, Philip HS},
  booktitle={Proceedings of the IEEE international conference on computer vision},
  pages={1529--1537},
  year={2015}
}

@inproceedings{unet+,
  title={An exploration of 2D and 3D deep learning techniques for cardiac MR image segmentation},
  author={Baumgartner, Christian F and Koch, Lisa M and Pollefeys, Marc and Konukoglu, Ender},
  booktitle={International Workshop on Statistical Atlases and Computational Models of the Heart},
  pages={111--119},
  year={2017},
  organization={Springer}
}

@article{mixup,
  title={mixup: Beyond empirical risk minimization},
  author={Zhang, Hongyi and Cisse, Moustapha and Dauphin, Yann N and Lopez-Paz, David},
  journal={arXiv preprint arXiv:1710.09412},
  year={2017}
}

@article{cutout,
  title={Improved regularization of convolutional neural networks with cutout},
  author={DeVries, Terrance and Taylor, Graham W},
  journal={arXiv preprint arXiv:1708.04552},
  year={2017}
}

@inproceedings{cutmix,
  title={Cutmix: Regularization strategy to train strong classifiers with localizable features},
  author={Yun, Sangdoo and Han, Dongyoon and Oh, Seong Joon and Chun, Sanghyuk and Choe, Junsuk and Yoo, Youngjoon},
  booktitle={Proceedings of the IEEE/CVF international conference on computer vision},
  pages={6023--6032},
  year={2019}
}

@inproceedings{puzzlemix,
  title={Puzzle mix: Exploiting saliency and local statistics for optimal mixup},
  author={Kim, Jang-Hyun and Choo, Wonho and Song, Hyun Oh},
  booktitle={International Conference on Machine Learning},
  pages={5275--5285},
  year={2020},
  organization={PMLR}
}

@article{comixup,
  title={Co-mixup: Saliency guided joint mixup with supermodular diversity},
  author={Kim, Jang-Hyun and Choo, Wonho and Jeong, Hosan and Song, Hyun Oh},
  journal={arXiv preprint arXiv:2102.03065
        
        },
  year={2021}
}

@article{acdc,
  title={Deep learning techniques for automatic MRI cardiac multi-structures segmentation and diagnosis: is the problem solved?},
  author={Bernard, Olivier and Lalande, Alain and Zotti, Clement and Cervenansky, Frederick and Yang, Xin and Heng, Pheng-Ann and Cetin, Irem and Lekadir, Karim and Camara, Oscar and Ballester, Miguel Angel Gonzalez and others},
  journal={IEEE transactions on medical imaging},
  volume={37},
  number={11},
  pages={2514--2525},
  year={2018},
  publisher={ieee}
}

@article{mscmr1,
  title={Multivariate mixture model for myocardial segmentation combining multi-source images},
  author={Zhuang, Xiahai},
  journal={IEEE transactions on pattern analysis and machine intelligence},
  volume={41},
  number={12},
  pages={2933--2946},
  year={2018},
  publisher={IEEE}
}

@inproceedings{mscmr2,
  title={Multivariate mixture model for cardiac segmentation from multi-sequence MRI},
  author={Zhuang, Xiahai},
  booktitle={International Conference on Medical Image Computing and Computer-Assisted Intervention},
  pages={581--588},
  year={2016},
  organization={Springer}
}

@InProceedings{luo2021urpc,
author={Luo, Xiangde and Liao, Wenjun and Chen, Jieneng and Song, Tao and Chen, Yinan and Zhang, Shichuan and Chen, Nianyong and Wang, Guotai and Zhang, Shaoting},
title={Efficient Semi-supervised Gross Target Volume of Nasopharyngeal Carcinoma Segmentation via Uncertainty Rectified Pyramid Consistency},
booktitle={Medical Image Computing and Computer Assisted Intervention -- MICCAI 2021},
year={2021},
pages={318--329}}

@inproceedings{wang2018video,
  title={Video captioning via hierarchical reinforcement learning},
  author={Wang, Xin and Chen, Wenhu and Wu, Jiawei and Wang, Yuan-Fang and Wang, William Yang},
  booktitle={Proceedings of the IEEE conference on computer vision and pattern recognition},
  pages={4213--4222},
  year={2018}
}

@inproceedings{lu2018neural,
  title={Neural baby talk},
  author={Lu, Jiasen and Yang, Jianwei and Batra, Dhruv and Parikh, Devi},
  booktitle={Proceedings of the IEEE conference on computer vision and pattern recognition},
  pages={7219--7228},
  year={2018}
}

@inproceedings{hu2020iterative,
  title={Iterative answer prediction with pointer-augmented multimodal transformers for textvqa},
  author={Hu, Ronghang and Singh, Amanpreet and Darrell, Trevor and Rohrbach, Marcus},
  booktitle={Proceedings of the IEEE/CVF Conference on Computer Vision and Pattern Recognition},
  pages={9992--10002},
  year={2020}
}

@inproceedings{marino2019ok,
  title={Ok-vqa: A visual question answering benchmark requiring external knowledge},
  author={Marino, Kenneth and Rastegari, Mohammad and Farhadi, Ali and Mottaghi, Roozbeh},
  booktitle={Proceedings of the IEEE/cvf conference on computer vision and pattern recognition},
  pages={3195--3204},
  year={2019}
}

@inproceedings{wang2016learning,
  title={Learning deep structure-preserving image-text embeddings},
  author={Wang, Liwei and Li, Yin and Lazebnik, Svetlana},
  booktitle={Proceedings of the IEEE conference on computer vision and pattern recognition},
  pages={5005--5013},
  year={2016}
}

@inproceedings{rombach2022high,
  title={High-resolution image synthesis with latent diffusion models},
  author={Rombach, Robin and Blattmann, Andreas and Lorenz, Dominik and Esser, Patrick and Ommer, Bj{\"o}rn},
  booktitle={Proceedings of the IEEE/CVF Conference on Computer Vision and Pattern Recognition},
  pages={10684--10695},
  year={2022}
}

@article{xue2021multi,
  title={Multi-modal co-learning for liver lesion segmentation on PET-CT images},
  author={Xue, Zhongliang and Li, Ping and Zhang, Liang and Lu, Xiaoyuan and Zhu, Guangming and Shen, Peiyi and Shah, Syed Afaq Ali and Bennamoun, Mohammed},
  journal={IEEE Transactions on Medical Imaging},
  volume={40},
  number={12},
  pages={3531--3542},
  year={2021},
  publisher={IEEE}
}

@article{fu2021multimodal,
  title={Multimodal spatial attention module for targeting multimodal PET-CT lung tumor segmentation},
  author={Fu, Xiaohang and Bi, Lei and Kumar, Ashnil and Fulham, Michael and Kim, Jinman},
  journal={IEEE Journal of Biomedical and Health Informatics},
  volume={25},
  number={9},
  pages={3507--3516},
  year={2021},
  publisher={IEEE}
}

@inproceedings{radford2021learning,
  title={Learning transferable visual models from natural language supervision},
  author={Radford, Alec and Kim, Jong Wook and Hallacy, Chris and Ramesh, Aditya and Goh, Gabriel and Agarwal, Sandhini and Sastry, Girish and Askell, Amanda and Mishkin, Pamela and Clark, Jack and others},
  booktitle={International conference on machine learning},
  pages={8748--8763},
  year={2021},
  organization={PMLR}
}

@inproceedings{huang2021gloria,
  title={Gloria: A multimodal global-local representation learning framework for label-efficient medical image recognition},
  author={Huang, Shih-Cheng and Shen, Liyue and Lungren, Matthew P and Yeung, Serena},
  booktitle={Proceedings of the IEEE/CVF International Conference on Computer Vision},
  pages={3942--3951},
  year={2021}
}

@article{zhou2022generalized,
  title={Generalized radiograph representation learning via cross-supervision between images and free-text radiology reports},
  author={Zhou, Hong-Yu and Chen, Xiaoyu and Zhang, Yinghao and Luo, Ruibang and Wang, Liansheng and Yu, Yizhou},
  journal={Nature Machine Intelligence},
  volume={4},
  number={1},
  pages={32--40},
  year={2022},
  publisher={Nature Publishing Group UK London}
}

@inproceedings{ronneberger2015u,
  title={U-net: Convolutional networks for biomedical image segmentation},
  author={Ronneberger, Olaf and Fischer, Philipp and Brox, Thomas},
  booktitle={Medical Image Computing and Computer-Assisted Intervention--MICCAI 2015: 18th International Conference, Munich, Germany, October 5-9, 2015, Proceedings, Part III 18},
  pages={234--241},
  year={2015},
  organization={Springer}
}

@article{Clark2013Cancer,
  title={The Cancer Imaging Archive (TCIA): maintaining and operating a public information repository},
  author={Clark, Kenneth and Vendt, Bruce and Smith, Kirk and Freymann, John and Kirby, Justin and Koppel, Paul and Moore, Stephen and Phillips, Stanley and Maffitt, David and Pringle, Michael and others},
  journal={Journal of digital imaging},
  volume={26},
  pages={1045--1057},
  year={2013},
  publisher={Springer}
}

@article{isensee2021nnu,
  title={nnU-Net: a self-configuring method for deep learning-based biomedical image segmentation},
  author={Isensee, Fabian and Jaeger, Paul F and Kohl, Simon AA and Petersen, Jens and Maier-Hein, Klaus H},
  journal={Nature methods},
  volume={18},
  number={2},
  pages={203--211},
  year={2021},
  publisher={Nature Publishing Group US New York}
}

@inproceedings{milletari2016v,
  title={V-net: Fully convolutional neural networks for volumetric medical image segmentation},
  author={Milletari, Fausto and Navab, Nassir and Ahmadi, Seyed-Ahmad},
  booktitle={2016 fourth international conference on 3D vision (3DV)},
  pages={565--571},
  year={2016},
  organization={Ieee}
}

@article{kirillov2023segment,
  title={Segment anything},
  author={Kirillov, Alexander and Mintun, Eric and Ravi, Nikhila and Mao, Hanzi and Rolland, Chloe and Gustafson, Laura and Xiao, Tete and Whitehead, Spencer and Berg, Alexander C and Lo, Wan-Yen and others},
  journal={arXiv preprint arXiv:2304.02643},
  year={2023}
}

@article{monteiro2018conditional,
  title={Conditional random fields as recurrent neural networks for 3d medical imaging segmentation},
  author={Monteiro, Miguel and Figueiredo, M{\'a}rio AT and Oliveira, Arlindo L},
  journal={arXiv preprint arXiv:1807.07464},
  year={2018}
}

@article{badrinarayanan2017segnet,
  title={Segnet: A deep convolutional encoder-decoder architecture for image segmentation},
  author={Badrinarayanan, Vijay and Kendall, Alex and Cipolla, Roberto},
  journal={IEEE transactions on pattern analysis and machine intelligence},
  volume={39},
  number={12},
  pages={2481--2495},
  year={2017},
  publisher={IEEE}
}

@InProceedings{Hatamizadeh_2022_WACV,
    author    = {Hatamizadeh, Ali and Tang, Yucheng and Nath, Vishwesh and Yang, Dong and Myronenko, Andriy and Landman, Bennett and Roth, Holger R. and Xu, Daguang},
    title     = {UNETR: Transformers for 3D Medical Image Segmentation},
    booktitle = {Proceedings of the IEEE/CVF Winter Conference on Applications of Computer Vision (WACV)},
    month     = {January},
    year      = {2022},
    pages     = {574-584}
}

@InProceedings{swinunetr,
    author    = {Tang, Yucheng and Yang, Dong and Li, Wenqi and Roth, Holger R. and Landman, Bennett and Xu, Daguang and Nath, Vishwesh and Hatamizadeh, Ali},
    title     = {Self-Supervised Pre-Training of Swin Transformers for 3D Medical Image Analysis},
    booktitle = {Proceedings of the IEEE/CVF Conference on Computer Vision and Pattern Recognition (CVPR)},
    month     = {June},
    year      = {2022},
    pages     = {20730-20740}
}

@article{chen2021transunet,
  title={Transunet: Transformers make strong encoders for medical image segmentation},
  author={Chen, Jieneng and Lu, Yongyi and Yu, Qihang and Luo, Xiangde and Adeli, Ehsan and Wang, Yan and Lu, Le and Yuille, Alan L and Zhou, Yuyin},
  journal={arXiv preprint arXiv:2102.04306
        
        },
  year={2021}
}

@article{swinunet,
  title={Swin-unet: Unet-like pure transformer for medical image segmentation},
  author={Cao, Hu and Wang, Yueyue and Chen, Joy and Jiang, Dongsheng and Zhang, Xiaopeng and Tian, Qi and Wang, Manning},
  journal={arXiv preprint arXiv:2105.05537
        
        },
  year={2021}
}

@inproceedings{lin2016scribblesup,
  title={Scribblesup: Scribble-supervised convolutional networks for semantic segmentation},
  author={Lin, Di and Dai, Jifeng and Jia, Jiaya and He, Kaiming and Sun, Jian},
  booktitle={Proceedings of the IEEE conference on computer vision and pattern recognition},
  pages={3159--3167},
  year={2016}
}

@incollection{can2018learning,
  title={Learning to segment medical images with scribble-supervision alone},
  author={Can, Yigit B and Chaitanya, Krishna and Mustafa, Basil and Koch, Lisa M and Konukoglu, Ender and Baumgartner, Christian F},
  booktitle={Deep Learning in Medical Image Analysis and Multimodal Learning for Clinical Decision Support},
  pages={236--244},
  year={2018},
  publisher={Springer}
}

@article{postdae,
  title={Post-DAE: anatomically plausible segmentation via post-processing with denoising autoencoders},
  author={Larrazabal, Agostina J and Mart{\'\i}nez, C{\'e}sar and Glocker, Ben and Ferrante, Enzo},
  journal={IEEE transactions on medical imaging},
  volume={39},
  number={12},
  pages={3813--3820},
  year={2020},
  publisher={IEEE}
}

@article{accl,
  title={ACCL: Adversarial constrained-CNN loss for weakly supervised medical image segmentation},
  author={Zhang, Pengyi and others},
  journal={arXiv preprint arXiv:2005.00328
        
        
        
        },
  year={2020}
}

@inproceedings{unet,
  title={U-net: Convolutional networks for biomedical image segmentation},
  author={Ronneberger, Olaf and Fischer, Philipp and Brox, Thomas},
  booktitle={International Conference on Medical image computing and computer-assisted intervention},
  pages={234--241},
  year={2015},
  organization={Springer}
}

@article{grady2006random,
  title={Random walks for image segmentation},
  author={Grady, Leo},
  journal={IEEE transactions on pattern analysis and machine intelligence},
  volume={28},
  number={11},
  pages={1768--1783},
  year={2006},
  publisher={IEEE}
}

@article{liu2022weakly,
  title={Weakly supervised segmentation of COVID19 infection with scribble annotation on CT images},
  author={Liu, Xiaoming and Yuan, Quan and Gao, Yaozong and He, Kelei and Wang, Shuo and Tang, Xiao and Tang, Jinshan and Shen, Dinggang},
  journal={Pattern recognition},
  volume={122},
  pages={108341},
  year={2022},
  publisher={Elsevier}
}

@inproceedings{S2L,
  title={Scribble2label: Scribble-supervised cell segmentation via self-generating pseudo-labels with consistency},
  author={Lee, Hyeonsoo and Jeong, Won-Ki},
  booktitle={Medical Image Computing and Computer Assisted Intervention--MICCAI 2020: 23rd International Conference, Lima, Peru, October 4--8, 2020, Proceedings, Part I 23},
  pages={14--23},
  year={2020},
  organization={Springer}
}

@inproceedings{qiu2023cor,
  title     = {CorSegRec: A Topology-Preserving Scheme for Extracting Fully-Connected Coronary Arteries from CT Angiography},
  author    = {Qiu, Yehui and Li, Zihan and Wang, Yining and Dong, Pei and Wu, Dijia and Yang, Xinnian and Hong, Qingqi and Shen, Dinggang},
  booktitle = {MICCAI},
  year      = {2023}
}

@article{kim2019mumford,
  title={Mumford--Shah loss functional for image segmentation with deep learning},
  author={Kim, Boah and Ye, Jong Chul},
  journal={IEEE Transactions on Image Processing},
  volume={29},
  pages={1856--1866},
  year={2019},
  publisher={IEEE}
}

@article{EM,
  title={Semi-supervised learning by entropy minimization},
  author={Grandvalet, Yves and Bengio, Yoshua},
  journal={Advances in neural information processing systems},
  volume={17},
  year={2004}
}

@inproceedings{li2022tfcns,
  title={TFCNs: A CNN-Transformer Hybrid Network for Medical Image Segmentation},
  author={Li, Zihan and Li, Dihan and Xu, Cangbai and Wang, Weice and Hong, Qingqi and Li, Qingde and Tian, Jie},
  booktitle={Artificial Neural Networks and Machine Learning--ICANN 2022: 31st International Conference on Artificial Neural Networks, Bristol, UK, September 6--9, 2022, Proceedings; Part IV},
  pages={781--792},
  year={2022},
  organization={Springer}
}

@article{li2023lvit,
  title={Lvit: language meets vision transformer in medical image segmentation},
  author={Li, Zihan and Li, Yunxiang and Li, Qingde and Wang, Puyang and Guo, Dazhou and Lu, Le and Jin, Dakai and Zhang, You and Hong, Qingqi},
  journal={IEEE Transactions on Medical Imaging},
  year={2023},
  publisher={IEEE}
}

@article{xu2022mrdff,
  title={MRDFF: A deep forest based framework for CT whole heart segmentation},
  author={Xu, Fei and Lin, Lingli and Li, Zihan and Hong, Qingqi and Liu, Kunhong and Wu, Qingqiang and Li, Qingde and Zheng, Yinhuan and Tian, Jie},
  journal={Methods},
  volume={208},
  pages={48--58},
  year={2022},
  publisher={Elsevier}
}

@inproceedings{shan2023coarse,
  title={Coarse-to-Fine Covid-19 Segmentation via Vision-Language Alignment},
  author={Shan, Dandan and Li, Zihan and Chen, Wentao and Li, Qingde and Tian, Jie and Hong, Qingqi},
  booktitle={ICASSP 2023-2023 IEEE International Conference on Acoustics, Speech and Signal Processing (ICASSP)},
  pages={1--5},
  year={2023},
  organization={IEEE}
}

@inproceedings{cciccek20163d,
  title={3D U-Net: learning dense volumetric segmentation from sparse annotation},
  author={{\c{C}}i{\c{c}}ek, {\"O}zg{\"u}n and Abdulkadir, Ahmed and Lienkamp, Soeren S and Brox, Thomas and Ronneberger, Olaf},
  booktitle={Medical Image Computing and Computer-Assisted Intervention--MICCAI 2016: 19th International Conference, Athens, Greece, October 17-21, 2016, Proceedings, Part II 19},
  pages={424--432},
  year={2016},
  organization={Springer}
}

@inproceedings{Yang2023NeRCo,
  title={Implicit Neural Representation for Cooperative Low-light Image Enhancement},
  author={Yang, Shuzhou and Ding, Moxuan and Wu, Yanmin and Li, Zihan and Zhang, Jian},
  booktitle={Proceedings of the IEEE/CVF International Conference on Computer Vision},
  year={2023}
}

@article{li2023chatdoctor,
  title={ChatDoctor: A Medical Chat Model Fine-Tuned on a Large Language Model Meta-AI (LLaMA) Using Medical Domain Knowledge},
  author={Li, Yunxiang and Li, Zihan and Zhang, Kai and Dan, Ruilong and Jiang, Steve and Zhang, You},
  journal={Cureus},
  volume={15},
  number={6},
  year={2023},
  publisher={Cureus}
}

@article{hong2023distance,
  title={A Distance Transformation Deep Forest Framework With Hybrid-Feature Fusion for CXR Image Classification},
  author={Hong, Qingqi and Lin, Lingli and Li, Zihan and Li, Qingde and Yao, Junfeng and Wu, Qingqiang and Liu, Kunhong and Tian, Jie},
  journal={IEEE Transactions on Neural Networks and Learning Systems},
  year={2023},
  publisher={IEEE}
}

@inproceedings{wang2023SwinMM,
  title     = {SwinMM: Masked Multi-view with Swin Transformers for 3D Medical Image Segmentation},
  author    = {Wang, Yiqing and Li, Zihan and Mei, Jieru and Wei, Zihao and Liu, Li and Wang, Chen and Sang, Shengtian and Yuille, Alan and Xie, Cihang and Zhou, Yuyin},
  booktitle = {MICCAI},
  year      = {2023}
}

@inproceedings{luo2022semi,
  title={Semi-supervised medical image segmentation via cross teaching between cnn and transformer},
  author={Luo, Xiangde and Hu, Minhao and Song, Tao and Wang, Guotai and Zhang, Shaoting},
  booktitle={International Conference on Medical Imaging with Deep Learning},
  pages={820--833},
  year={2022},
  organization={PMLR}
}
\end{document}